\documentclass[sigconf]{acmart}

\usepackage{graphicx}
\usepackage{hyperref} 

\renewcommand\footnotetextcopyrightpermission[1]{} 
\newlength{\halfwidth}
\begin{document}

\title{A Trillion Genetic Programming Instructions per Second}

\author{\href{http://www.cs.ucl.ac.uk/staff/W.Langdon}
{W. B. Langdon}%
}
\affiliation{%
  \institution{Department of Computer Science, University College London,
Gower Street,
London, UK}
  \streetaddress{Gower Street}
  \state{}
  \postcode{WC1E 6BT, UK}
}
\email{W.Langdon@cs.ucl.ac.uk}

\begin{abstract}
We summarise how a 3.0GHz 16 core AVX512 computer can interpret
the equivalent of up to on average 1\,103\,370\,000\,000 GPop/s.
Citations to existing publications are given.
Implementation stress is placed on both parallel computing,
bandwidth limits
and avoiding repeated calculation.
Information theory suggests in digital computing,
failed disruption propagation
gives huge speed ups as FDP and incremental evaluation can be used to
reduce fitness evaluation time in phenotypically converged populations.
Conversely FDP may be responsible for evolution stagnation.
So the wider Evolutionary Computing, Artificial Life, Unconventional Computing
and Software Engineering community
may need to avoid deep nesting.
\end{abstract}


\maketitle

\section{Cheats and how you can used them}

To come clean.
\begin{itemize}
\item
{\em Up to} $1.1\ 10^{12}$ GPop/s is from one run.
The median $0.65\ 10^{12}$ GPop/s is more typical
\cite[Tab.~3]{Langdon:2022:ALJ}.

\item
We used 16 threads on one node of a shared cluster
which supported 48 threads.
(The cluster job scheduler used the other 32 threads
for a wide mix of jobs during our runs.)

\item
The operating system manages the individual CPU core clocks
and is free to reduce them below 3.0GHz.

\item
This is the mean performance across the whole run
and so includes everything,
including gathering statistics.
(In this and another run
we turned off gathering some unused population statistics,
such as calculating the height of the trees.)

\item
{\em Equivalent of}.
This is the one that matters.
The headline figure is based on assuming each GP opcode
is interpreted on each test case.
We really do use every test case every time but the big
result is to avoid repeated work,
Section~\ref{sec:avoidwork}.

\hspace{1ex}
Each generation is created from its parent generation.
We have evaluated all of the parents.
Their children are only marginally different.
We do our best to inherit their fitness evaluation
within the constraints of not using more memory.

\end{itemize}

\section{Background}

My goal is summarise recent improvement in genetic programming
implementations in the hope that the ideas used may be beneficial
elsewhere. 
Section~\ref{sec:small_mem_ga} describes how to reduce
memory usage by more than 50\%,
whilst
Section~\ref{sec:par} is concerned with exploiting parallel hardware.
In converged populations of enormous trees
incremental fitness evaluation~\cite{langdon:2021:EuroGP}
(Section~\ref{sec:avoidwork})
proved so successful at reducing fitness evaluation run time
that crossover became the dominant cost,
so Section~\ref{sec:fast_xo} describes ways
to reduce the cost of crossover.
The success of incremental evaluation also lead to
consideration of information flows within GP trees
and more widely in hierarchical structures,
inspiring failed disruption propagation, FDP~\cite{Petke:2021:FSE-IVR},
and the realisation that deep nested structures will be
resistant to evolution and,
as discussed in Section~\ref{sec:FDP},
the suggestion that to evolve large complex artefacts,
they will have to be composed of many shallow
sub-modules~\cite{langdon:2021:sigevolution}
(see Figure~\ref{fig:lung_like}, 
   page~\pageref{fig:lung_like}).
Details can be found in the published papers and C++~code.

\subsection{Faster Genetic Programming}

Genetic Programming~\cite{koza:book},
like many Artificial Intelligence techniques based on search,
can be expensive in terms of computer resources.
Indeed Koza in the first GP book~\cite{koza:book}
describes ways to speed it up~\cite{poli08:fieldguide}.

Until recently (Section~\ref{sec:fast_xo}),
as is common with evolutionary computing in general,
almost the whole GP runtime is taken by
the time to do fitness evaluation
and so speed up techniques have concentrated upon
faster fitness functions.

\subsection{Faster Fitness}
\label{sec:fasterfit}

Almost all GP (as we do) interprets the evolved code.
However, particularly when GP individuals are run many times,
e.g.\ because there are many test cases,
it can be beneficial not to interpret the code
but to compile it
and run the compiled binary code
(e.g.\ Harding~\cite{eurogp07:harding},
Gregor and Spalek~\cite{Gregor:2016:ELEKTRO}).
However running a traditional compiler is expensive
and 
\label{p.Fukunaga}
Fukunaga et al.~\cite{fukunaga:1998:gchpGP} and 
Vasicek and Slany~\cite{vasicek:2012:EuroGP} 
have advocated generating
machine code inside the GP system
and executing that.
Indeed Nordin~\cite{Nordin:1995:tcp,nordin:thesis}
evolved executable machine code directly,
leading to Discipulus~\cite{francone:manual}
which evolved Intel x86 binary machine code~%
\cite{foster:2001:discipulus}.

\subsection{Better Fitness}

Fundamentally the goal of the fitness function is not
to measure how well an evolved individual works
or solves a problem but to guide search
and essentially fitness and selection boils
down to a single bit:
does this individual get a child or not.
The evolutionary computing community,
puts a huge amount of effort into the fitness function,
whose output is often corrupted by the addition of 
stochastic selection noise.
This single (noisy) bit is often extremely expensive.

Many techniques have been advanced to reduce the overall cost
of the fitness function and/or provide better results.
Sometimes the goal is to reduce costs but
(also) it may be to drive evolution in a different direction.
 
Recognising that the child/no child decision is often made
using tournament selection,
Teller and Andre~\cite{Teller:1997:acnfc} devised a scheme
which aimed at using statistical tests whereby,
assuming a random sample,
only enough tests are run by the fitness function to
reliably tell which of a candidate set of parents
would win the tournament if all the tests were run.
That is, their ``rational allocation of trials'' (RAT)
aims (although with some noise)
to drive evolution as before but with considerably
less effort.

Treating individuals with zero children as pure overhead to the GP system,
Tackett proposed ``soft brood selection''
\cite{Tackett:1994:broodGP},
which aims to dispose of likely unfit children
without going to the expense of growing them all to adulthood.

When evolving agents~\cite{langdon:1998:antspace}
or programs~\cite{Langdon:2012:mendel,Petke:gisurvey},
an individual which enters a never ending loop will have low fitness
but high fitness evaluation cost.
To reduce fitness effort,
it is common in such cases to impose a timeout
and cut short fitness evaluation
when the imposed time limit is reached.
However the chosen limit is application specific.
(This also has the unfortunate effect
that when different limits are chosen for the same ``benchmark''
\cite{langdon:1998:antspace}
subtle difference arise
between supposedly the same system.)

To try and reduce the impact of arbitrary time limits,
Maxwell proposed a ``coroutine'' model~\cite{icec94:maxwell}
in which (like a conventional operating system time sharing system),
each member of the population gets a quantum of computer time 
to run fitness tests.
Where selection,
i.e.~child/no child, is not clear cut,
aspiring individuals may get additional quanta to try and assemble more fitness points.
Again the goal is to avoid ``wasting'' fitness effort,
but accepting the introduced noise may drive evolution in another direction.

Gathercole~\cite{ga94aGathercole,gathercole:thesis}
proposed schemes for dynamically changing which 
fitness tests are used as evolution proceeds.
Gathercole's DSS scheme was used commercially
\cite{francone:manual}.

Spector's Lexicase~%
\cite{Spector:2012:GECCOcompA}
can be thought of having elements of both RAT and DSS\@.
Again it runs only a varying subset of the available fitness test cases
but its primary motivation is to 
evolve a better solution, rather than faster fitness evaluation.

Some of these schemes have elements of 
coevolution~\cite{Reynolds:1994:tag}
about them.
In which the fitness test cases are thought of as a separate
population and the primary goal is better evolution
rather than reduced fitness effort.

\subsection{\mbox{Distributed Populations: Slower Convergence}}

The implementation changes to GPquick
(Sections~\ref{sec:small_mem_ga} to~\ref{sec:avoidwork})
allow the study of the evolution of far bigger
genetic programming systems
within the constraints of the available computers.
The implementation takes pains to sequester
the use of pseudo random operations,
and synchronise parallel operations,
so that the results are identical
to those before the improvement were made
and experiments can be repeated.

By relaxing the exact reproducibility constraint you can
ease the implementation of
exciting stochastic effects; 
such as studying immigration/emigration 
between demes and island sub-populations.
These may not only easily allow operation over
multiple computers (Section~\ref{sec:par})
but also delay convergence~%
\cite{langdon:book},
\cite{DBLP:journals/heuristics/AffenzellerW04}.

\section{\mbox{Generational $\le$\! Steady State Memory}} 
\label{sec:small_mem_ga}

Although introduced for the benefit of genetic programming
populations with large trees,
these algorithms apply to any evolutionary algorithm~(EA) with
a fixed population size~$M$.
For example, they might be useful for
very large traveling salesman problems (TSP),
such as Drori et al.'s proposed billion city 
Galaxy TSP benchmark~\cite{drori2018galaxytsp}.

Long ago 
Koza et al.~\cite[pages 1044-1045]{koza:gp3}
showed that a generational evolutionary algorithm need
take no more memory than a steady state EA\@.
Essentially at the start of each generation
during crossover and mutation they maintain multiple
queues of parents according to the number of children
they have yet to create.
As each crossover or mutation is performed,
the parents' number of unborn children are reduced
and when necessary the parents are moved between queues.
Parents with no yet to be created children
are deleted immediately,
allowing their buffers to be used by new children.
The queue of parents with only one unborn child
is dealt with first.
If it is empty, the queue holding parents
with two children waiting to be created is processed.
Of course processing the first item in this queue,
may result in moving individuals
to the first queue.
If so, it will again be processed first.

Koza et al.\ allow two parent crossover to create two children.
Their queues can be slightly simplified if each
genetic operation only generates one child
\cite{langdon:2020:small_mem_ga_2pages}.
The essential idea remains;
process those individuals which will free resources first.

I have not done this,
but the individual parents within each queue
could be ordered to process 
those that will free most memory first.
(Note the current implementation uses un-ordered queues
so that each operation takes constant time
and the overhead is more-or-less the same as before.)
Alternatively,
to balance the threads
in a multi-threaded environment,
you might want
to start processing operations that will take the longest time first.
Unlike incremental fitness evaluation~\cite{langdon:2021:EuroGP},
the time taken by crossover or mutation can be accurately estimated
in advance from the size of the crossover fragments to be moved
and the computer's memory bandwidth.

Like my earlier implementations of GPquick~\cite{singleton:byte},
there are many
generational EA systems which 
for a population of $M$~individuals
simply create the new generation whilst keeping the old
one and thus need twice as much memory (i.e.~$2M$ buffers).
Using a single threaded version of Koza et al.\  scheme
and two parent crossover
a single threaded implementations needs at most $M+2$
buffers.

As my trees became bigger,
it became important to move GPquick from using
$2M$ buffers.
Indeed as fitness evaluation got faster 
\cite{langdon:2019:gpquick,langdon:2020:cec}
it became worthwhile running crossover
(like fitness evaluation)
in parallel.
With synchronisation locks the child birthing
queues can be processed in parallel.
Notice the order they are processed in now depends 
upon the exact thread timings and will typically vary between runs.
However all the tasks are processed and the results
are the same.
GPquick generates one child per crossover.
The parallel Linux pthreads~($t$) version needs at most
$M+2t$ buffers
\cite{langdon:2020:small_mem_ga_2pages,langdon:2020:small_mem_ga}.

If low fitness individuals are discarded,
with two parent crossover,
in practise
the memory can be reduce 
to about $0.65M$.
With some virtual memory systems,
it may be possible to allocate $M+2t$ buffers
but in most generations actually only use two thirds of the memory.

Reducing memory usage below $M$ buffers
has become possible
due to
innovations introduced to reduce the cost of 
memory buffer copies in crossover.
Such as 
evaluating fitness before crossover
(Section~\ref{sec:fitness_first})
and extracting and separately storing
the small subtrees to be donated to children 
in the next generation 
\cite{Langdon:2021:GECCO,Langdon:2021:GPTP}
(also Section~\ref{sec:fatherless}).

\subsection{GPquick uses Flatten Linear GP Trees}

For efficiency~\cite{kinnear:keith},
GPquick~\cite{singleton:byte} flattens each GP tree
into a linear array of opcodes.
Reproduction, mutation, crossover and fitness evaluation
all use this.
For work on parallel graphics accelerators
(GPUs)~\cite{langdon:2008:eurogp}
(see Section~\ref{sec:gpu}),
the normal depth first tree order can be replaced
by reverse polish notation (RPN),
which processes in left to right order the RPN expression,
explicitly pushing intermediate results on to a stack,
rather than recursively processing the tree (with an implicit stack).
In both cases the depth of the stack used is the tree height.
Again reproduction, mutation, crossover and fitness evaluation
all use the linear buffer,
avoiding the need for data conversion between the CPU and GPU\@.

To avoid heap fragmentation GPquick
allocates a user defined number of fixed sized byte arrays
to hold the GP population at the start of the run.
Usually the flatten evolved tree does not fill the whole of its
buffer,
in which case the contents of the rest of the buffer is left
undefined.

For applications with many test cases,
GPquick has a FASTEVAL compile time option.
If FASTEVAL is enabled,
a single pass through each GP tree is inserted before fitness
evaluation,
which replaces each byte sized opcode by a pointer
(occupying 8 bytes) 
to the function inside the fitness evaluation code responsible
for evaluating the opcode.
For every test case,
this avoids a table look up to translate each opcode to 
a function pointer.
However, for simplicity and to save memory,
I did not use FASTEVAL\@.
Notice the FASTEVAL pass is analogous to
the translation from GP tree to 
binary machine code advocated by
Alex Fukunaga and others mentioned above 
in Section~\ref{sec:fasterfit}.

\section{Parallel Evaluation}
\label{sec:par}

It has long been recognised that GP in particular
and evolutionary computing in general,
are well able to take advantage of parallel computing,
e.g., \cite{Fraser:1994:inkbiro},
\cite{juille:1995:fgSIMD,pollack:1996:aigp2},
\cite{Koza:1995:pGPnt},\cite{bennett:1999:BPCSPHPD},
\cite{PDPTA96b},
\cite{bolis:2001:EuroGP,Maitre:2013:ecgpu}, 
\cite{DBLP:conf/ppsn/ArenasCEJGPPS02}, 
\cite{Cahon:2004:JoH,DBLP:journals/heuristics/AffenzellerW04},
\cite{harding_genetic_2008,langdon:2008:eurogp,langdon:2008:SC},
\cite{FernandezdeVega:pdci}, 
\cite{langdon_2012_sigevolution},
\cite{Gong:2015:ASC}, 
\cite{Burlacu:2020:GECCOcomp}, 
\cite{LaCava:2021:NeurIPS}. 

\subsection{Parallel Threads and Multi-Core CPUs}
\label{sec:pthreads}

The previous section has already mentioned 
the use of Linux pthreads to perform crossover in parallel.
In fact multi-threading was implemented in GPquick
to speed up fitness evaluation
\cite{langdon:2019:gpquick}.

With a population that is bigger than the number of cores
simply giving the next operation
(e.g.\ child to create
or fitness to evaluate)
to each thread as it finishes its previous task
(with appropriate synchronisation locks)
works well and usually shares the work equitably between
the available compute cores.

If the tasks are very unequal,
this simple way of handing out work may be problematic.
For example,
with incremental evaluation
and huge trees,
it was sometimes the case that one fitness evaluation
took far longer than the others,
causing the whole population to wait
for one fitness evaluation to complete.
I have not investigated better ways to share tasks between threads,
but an easy alternative might be simply to run
more independent jobs on your computer than it has cores.
With our Centos version of Linux,
the operating system overhead of rescheduling work on a core is low.
Therefore, if you have more than one EA task
and sufficient memory,
allowing the operating system to keep the cores
busy could be an efficient and easy to implement solution.

\subsection{Parallel Fitness Evaluations using
SIMD Vector Instruction CPUs}
\label{sec:avx}

Riccardo Poli pointed out in bit-wise problems
the basic hardware is capable of doing 32 or 64 bit operations
simultaneously
\cite{poli:1999:aigp3,poli:2000:22par}.
Indeed I used a 64~bit Intel computer to do all
64 fitness cases in parallel for 
the first long term evolution experiment (LTEE)
\cite{Langdon:2017:GECCO}
running GP on a binary tree version of Koza's~\cite{koza:book}
6-multiplexor problem
for 100\,000 generations%
\footnote{
Rich Lenski's team \cite{Lenski:2015:PRoySocB}
have been running microbiology laboratory evolutionary experiments
on the bacteria E.~coli since 1988~\cite{lenski1988},
which have now reached
73\,500 generations~\cite{Kokko:2021:rspb}.
}.
Some current Intel hardware supports 512 bit operations.
I am unaware of any genetic programming
work using AVX-512 to process 512 Boolean test cases in parallel.
However it appears
Hrbacek and Sekanina~\cite{Hrbacek:2014:GECCO}'s 
impressive work 
with cartesian genetic programming
using 256 bit SSE instructions could be readily 
extended to the newer 512 AVX.

In addition to bit-wise operations, the AVX vector instructions
can simultaneously perform 16 floating point calculations.
I wrote an AVX interpreter for GPquick
so that instead of processing each tree once per test case
it passed through each tree only once and
instead executed each opcode on all the test cases.
The first AVX version used the traditional depth first ordering
which requires each node within the tree to recursively
call the subtrees below it
and store intermediate results on a stack.
For efficiency an external stack
(of depth equal to the height of the tree)
is used.
GPquick requires the user to specify a maximum size of the trees
before it starts.
From the maximum size an estimate can be made of the maximum
height of any evolved tree
and used to set the size of the stack
when GPquick starts.

Vasicek and Slany \cite{vasicek:2012:EuroGP}, 
like Hrbacek and Sekanina~\cite{Hrbacek:2014:GECCO},
used 256 bit SSE instructions to allow 
cartesian genetic programming's fitness function to process
up to eight floating point numbers simultaneously.
Again it would seem possible to extend this to the newer 512 AVX
now available on some Intel servers
as well as Intel specialist plug-in GPU like accelerators.

De~Melo et al.~\cite{DeMelo:2020:evoapplications}
also used AVX parallel hardware and
claim an impressive median peak performance of 7.5 billion
genetic programming operations per second in a single CPU core,
however this is for their interpreter alone
and does not include the rest of their GP system.

\subsection{Alternative Parallel Approaches}

\subsubsection{Hybrid Computing, Compute Accelerators: GPU and TPU}
\label{sec:gpu}

Many flavours of hybrid architecture are possible.
Here we mean a non-X86 or Intel Xeon boards,
which may be housed inside the computer
and certainly within a few feet or it.
Although experiments continue with FPGA's~\cite{tgordon},
running computation on graphics cards (GPUs) and domestic gaming boards,
usually manufactured by nVidia
\cite{langdon:2008:eurogp},\cite{langdon:2013:ecgpu},
and known as general purpose computing on GPUs,
GPGPU~\cite{owens_2007_cgf},
continues to be popular.
However more recently,
especially for training deep neural networks,
Google's TPU
hardware accelerators have become popular.

Although perhaps for Pandemic related reasons,
there is perceived to be a worldwide shortage of silicon chips
and the steep increase in the GPU price performance curve
has soften,
it is expected that the impressive raw performance of
GPU and other accelerators will continue to rise.
Indeed already GPUs with multiple gigabytes of on board memory and
thousand of fully capable computer cores
are in use.
Although traditionally it is tricky to code GPGPU to get the best of the hardware:
evolutionary computing,
indeed genetic programming
\cite{langdon_2012_sigevolution},
have been shown to be fully capable of exploiting GPUs.

\subsubsection{Data Center Networks: MPI}
It is clear that CPU clock speeds are stuck at around 3~GHz
and no great speed up can be expected in that direction.
However compute power has continued to increase
by allowing greater use of parallel hardware.
Sections~\ref{sec:pthreads}
and~\ref{sec:avx}
have discussed multi-core and
SIMD vector instructions.
Further increases in both can be expected.
However the dominant growth in computer power
has been in loosely coupled
clusters of compute servers.
Indeed, today it is not uncommon for 
universities to run their own clusters of thousands of cores.
And academics have not been slow in exploiting
this architecture,
originally called Beowulf computing
\cite{bennett:1999:BPCSPHPD},
and nowadays referred to as cluster, grid~\cite{Wagner:2004:SS}
or cloud computing.
Indeed many use commercial off campus cloud computing,
e.g.\
Amazon  AWS EC2,
Google Cloud and Microsoft Azure.

Although many protocols exist for sharing data
between peer networks,
MPI%
\footnote{%
It seems that MPI has replaced the earlier PVM,
e.g., as used in the distributed evolutionary algorithm
ParadisEO~\cite{Cahon:2004:JoH}.}
is popular in the Linux world 
when interconnecting compute nodes within the same
data centre.

\subsubsection{Global Networks: HTTP \& P2P}

It has been known for a long time that
it is possible to distribute evolutionary computing 
globally~\cite{chong:1999:jDGPi,Laredo:2014:GPEM}.
Often this is done by distributing both population and
fitness evaluation,
for example Laredo et al.'s 
EvAg~\cite{Laredo:2009:GPEM}.
However perhaps the most famous example is
SETI@home~\cite{Anderson:2002:CACM},
which uses otherwise idle compute time donated by
volunteers world wide.

\begin{figure}
\includegraphics[scale=0.299287]{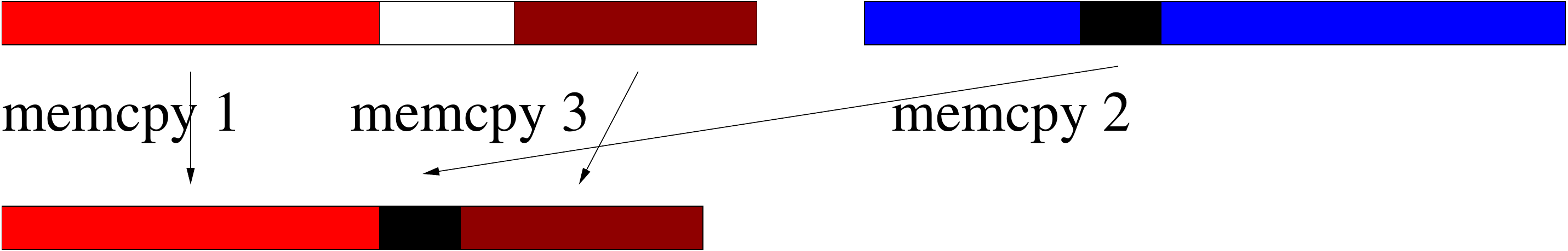}
\caption{\label{fig:crossover}
Original GPquick 
subtree crossover
requires three \mbox{memcpy} buffer copies:
1)~root segment of donating parent (mother, red/brown) is copied to offspring buffer.
2)~subtree from second parent (father, blue/black) is copied to offspring.
3)~tail (brown) of $1^{\rm st}$ parent copied to child.
}
\end{figure}

\begin{figure}
\flushleft
\includegraphics[scale=0.299287]{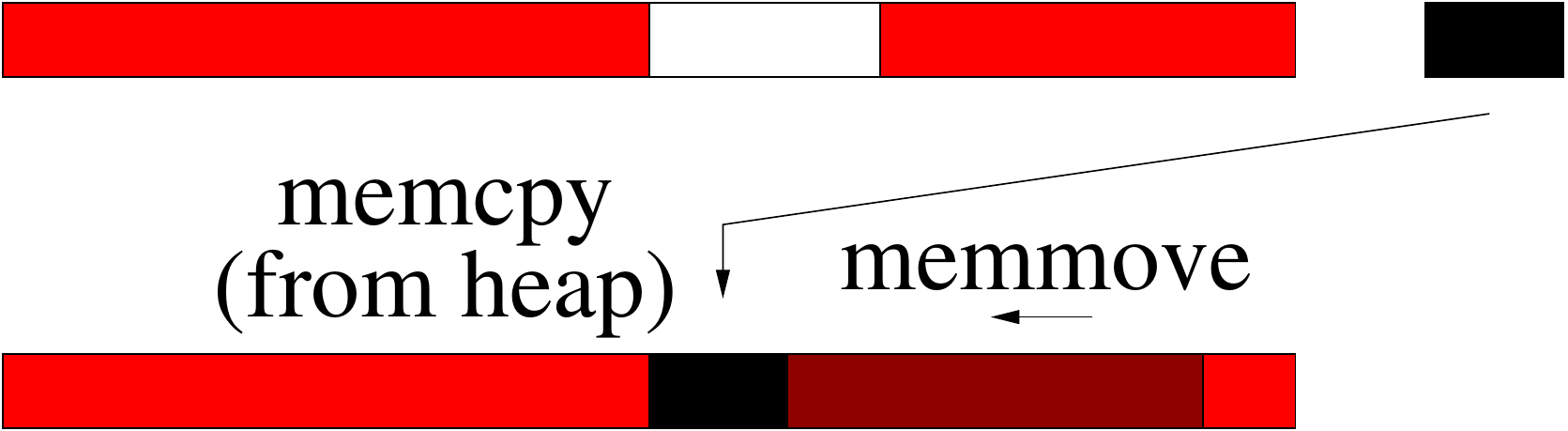}
\caption{\label{fig:crossover2}
In place subtree crossover uses only one GPquick buffer.
The offspring is the last child of the $1^{\rm st}$ parent (mother)
and reuses its buffer.
Another optimisation is that
the only part of the $2^{\rm nd}$ parent (father) to be kept is the
subtree to be inserted (black).
It is small enough to be stored on the heap.
In 71\% of children the 
subtree to be remove (white) and to be inserted (black)
are different sizes,
and so memmove is used to 
shuffle the second part of the mother's buffer (brown) up or down.
Finally the father subtree overwrites the buffer.
The unused part of the buffer (right end) is not cleared
and so may contain junk from earlier generations.
}
\end{figure}

\section{Faster Crossover}
\label{sec:fast_xo}

GPquick uses a very compact representation
for its GP trees.
This only requires one byte per tree node
\cite[pp~297]{kinnear:keith}.
Its compactness allows both very large populations
(e.g.\ 5~million trees~\cite{langdon:2008:SC})
and very large trees,
(e.g.\ two billion nodes~\cite{Langdon:2022:ALJ}).
It was only when fitness evaluation was sped up dramatically
that the cost of crossover became noticeable.
When the evolved trees become very large,
reproduction, e.g.\ crossover and mutation,
becomes expensive in GPquick
due to the volume of data to be copied,
see Figure~\ref{fig:crossover}.

GPquick now uses several ways to speed up crossover:
\begin{itemize}
\item
Run in parallel thread (see Section~\ref{sec:pthreads} above)

\item
Run fitness evaluation before genetic manipulation.
Thus children of fit parents who have
low fitness, or are just unlucky,
and so have no children themselves
need never be created
\cite{Langdon:2021:GECCO,Langdon:2021:GPTP}.

\item ``Fatherless'' crossover
\cite{Langdon:2021:GECCO,Langdon:2021:GPTP}
(see also Section~\ref{sec:fatherless}).

\hspace{1ex}
To allow exact reproduction,
GPquick performs
stochastic activities, such as choosing crossover points,
in a single master thread
before per generation parallel activities.
Since the crossover fragments are known,
it becomes easy to copy them all before
any genetic operations are done.
This means:
\begin{itemize}
\item Parents (``Fathers'') who only donate subtrees can be deleted
immediately.
Indeed only parents which donate their root node to at least one child
need be saved and processed by the crossover queues
(Section~\ref{sec:small_mem_ga}).
This saves full length buffers
and potentially reduces actual memory requirement below $M$~buffers.

\item
The copied subtree fragments are small enough to keep on the heap
and so potentially allow better use of the CPU's L2 and L3 caches.

\item
The crossover/mutation queues (Section~\ref{sec:small_mem_ga})
also make it easy to spot
when a parent has only one child left to be born.
In such cases, the new child is not allocated
a new buffer but simply reuses its parent's buffer.
(This is known as in place crossover.
Contrast Figure~\ref{fig:crossover2}
with Figure~\ref{fig:crossover},
both from~\cite{Langdon:2021:GECCO}.)
This will usually require about half the existing buffer to be
shuffled
but this is usually faster than copying almost the whole buffer
to a new location
\cite[Sec.~5]{Langdon:2021:GPTP}.

\end{itemize}

\end{itemize}

\noindent
GNU glibc-2.19's memmove drops into X86 assembler
to move opcodes in the buffer.
It uses a combination of 
{\tt rep~movsb} and
{\tt rep~movsl}
without the need to take a copy of the data.
The need to move the data up or down within the buffer arises when
the subtree removed by crossover is not the same size
as the subtree to be inserted
(Figure~\ref{fig:crossover2}).
Most crossover fragments are small
and so usually the buffer need be shuffled by two or perhaps four bytes.
With large trees,
memmove is on average 14\% faster than memcpy
\cite[Sec.~5]{Langdon:2021:GPTP}.
This may be because most of its
{\tt movsl} source and destinations are in the same L1 cache line
whereas with memcpy they reference different lines.

\section{Avoiding Fitness Evaluation Work}
\label{sec:avoidwork}

The best way to optimise anything is not to do the existing work better
but to reduce the amount of work to be done.
Section~\ref{sec:gpinc}
describes in fitness converged populations how to avoid evaluating
more than 99\% of opcodes 
using a method based on incremental fitness evaluation
\cite{langdon:2021:EuroGP}.

\subsection{Population as a directed acyclic graph}

Simon Handley~\cite{Handley:1994:DAGpcp}
had a very elegant GP representation 
in which the whole evolving population of genetic programming trees
was represented by a directed acyclic graph (DAG).
Each individual was represented by an entry point into the common code.
Common subtrees were held only once in the DAG\@.
When fit individuals were selected for crossover,
they exchanged existing subtrees.
The DAG kept track of this,
thus compactly representing the whole population.
See also Nic McPhee et al.'s Sutherland~\cite{mcphee:1998:sutherland}.

In the absence of side effects,
when the same fitness test cases are used in the next generation,
the evaluation of the each subtree is unchanged in the next generation
and the DAG can be used to cache subtree evaluations.
On many non-image processing tasks,
it is feasibly to cache every subtree evaluation.
Meaning after each genetic operation,
only the sub-evaluations from the crossover or mutation
site to the tree's root node need be re-evaluated.
That is,
fitness evaluation time
no longer scales as the size of the trees
but as their depth.
However...

Caching partial evaluations of the whole population
would defeat GPquick's compact opcode representation
(Section~\ref{sec:fast_xo})
and if trees are allowed to grow without restriction
it would not be feasible. 

\subsection{Bottom Up Fitness Evaluation}
\label{sec:gpinc}

What to do?
As is common in evolutionary computing,
we used the same test cases every generation.
The huge evolved trees were converging and thus
burning huge amounts of CPU time
to calculate the same fitness values over and over again.

In the {\bf absence of side effects} the GP trees are pure functional expressions
and thus can be evaluated in any order.
I.e.\ they do not have to be evaluated 
in the traditional root first recursive top down way.
Instead incremental fitness evaluation~\cite{langdon:2021:EuroGP},
evaluates each new child from the change point outwards.
Remember the only change between parent and child
is here.
All the rest of the code is identical
and, unless the code below it has changed,
it will evaluate identically both in the parent and its offspring.

\begin{figure}
\centerline{\includegraphics[scale=0.149884]{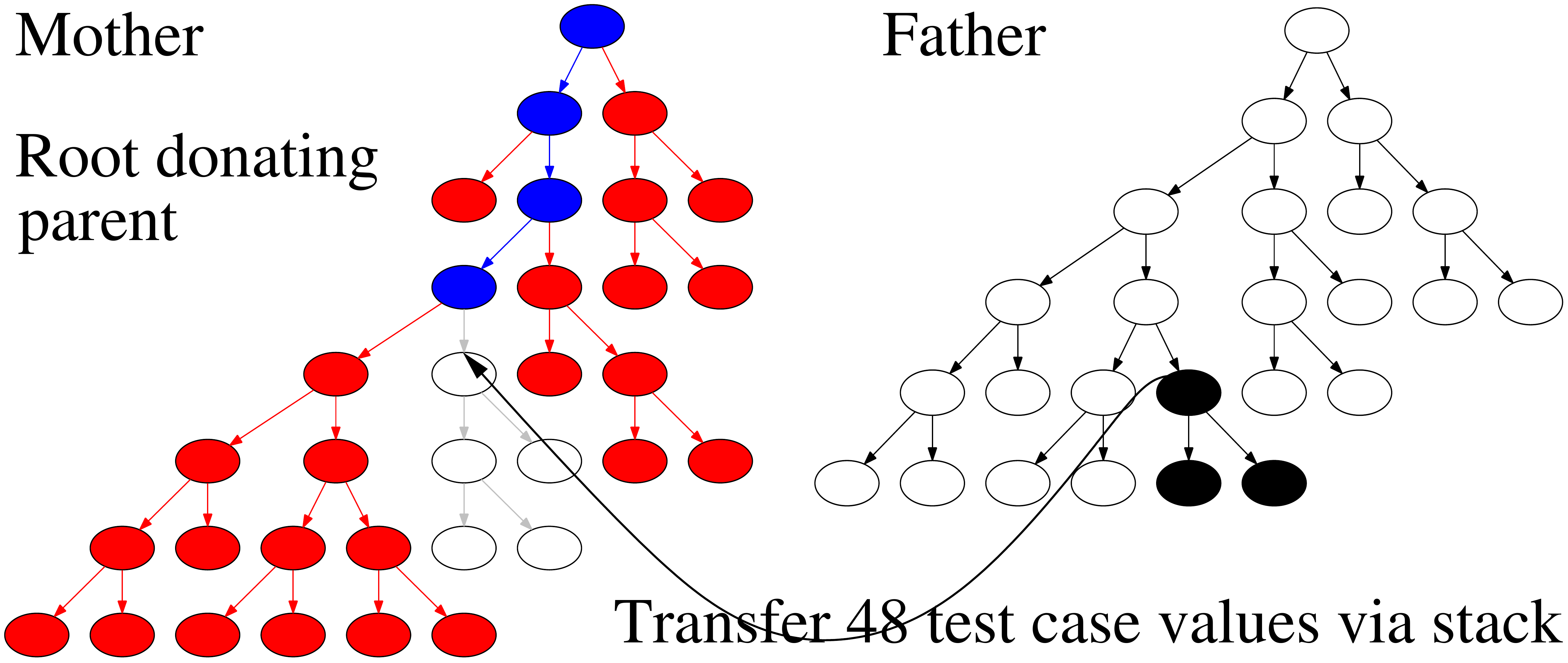}}
\caption{\label{fig:incfit}
With fitness first,
fitness is evaluated using only the parents,
i.e., before the child is created by crossover.
Assuming no side effects,
the subtree to be inserted (black) is evaluated on all test cases
and values are transferred to evaluation of mother at the location of
the subtree to be removed (white).
Using incremental evaluation~\protect\cite{langdon:2021:EuroGP}
differences between the original
code (white subtree) and the new, are propagated up the $1^{\rm st}$
parent (mother) until either all differences are zero
or we reach the root node.
Based on~\protect\cite{Langdon:2021:GECCO}.
}
\end{figure}

\begin{figure*}
\centerline{\includegraphics[width=\textwidth]{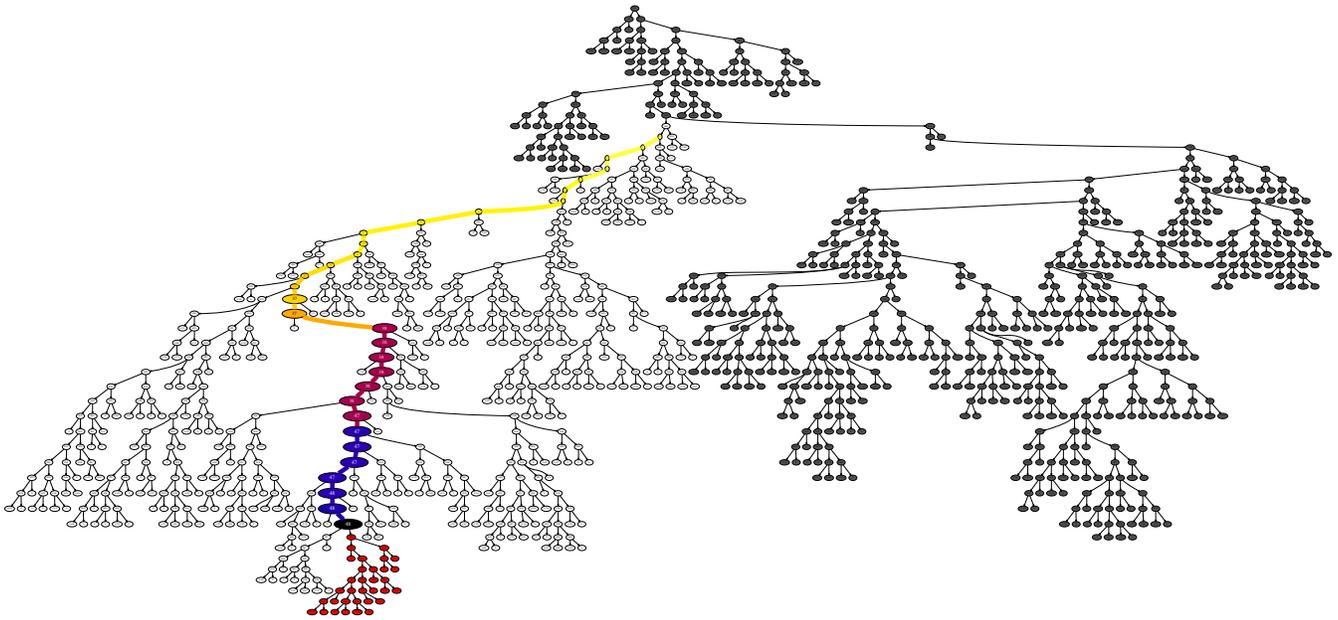}}
\caption{\label{fig:david2_colour}
The impact of a crossover (red subtree) dissipates
the further away from the crossover site we measure it.
The change in fitness test values is shown with coloured nodes.
The size of the coloured nodes shows the number of disrupted test cases.
The colour shows, on a log scale, the average difference
in evaluation on the remaining disrupted nodes.
Brightest yellow shows smallest non-zero difference
\mbox{(RMS $3.1\ 10^{-10}$)}. 
Eventually this crossover has no fitness impact at all.
White nodes are identical before and after the crossover.
Dark grey nodes show the part of the tree which does not have to be evaluated.
From~\protect\cite{langdon:2021:ieeeblog}.
}
\end{figure*}

With incremental evaluation,
GPquick
re-evaluates the ``mother''
(the parent who donates the root node to the child)
as it evaluates her child,
see Figure~\ref{fig:incfit}.
Firstly notice in almost all the tree 
the two evaluations are identical,
so the worst case overhead is modest.
Secondly the only place they can be different,
is on the chain of function calls from
the changed code to root node.
If at any point along that chain,
the evaluation of the new code
is equal to that in its parent,
then the two evaluation will
remain identical from that point to the root node.
If they are identical on all test cases,
the child's fitness will be identical
to that of its ``mother''.
Thus if at any point on the route up the tree
from the crossover or mutation site
to the root node
both evaluations on all the test cases
are identical,
the child's fitness is known to be
that of its parent and we can immediately stop
fitness evaluation.
 (Figure~\ref{fig:david2_colour},
page~\pageref{fig:david2_colour},
shows fitness disruption 
caused by a crossover fading a way in a small tree.)
In large converged populations
the savings can be two or three orders of magnitude.
In fact so fast,
that the cost of crossover becomes dominant.

\subsection{\mbox{Fitness Selection Before Crossover/Mutation}}
\label{sec:fitness_first}
\label{sec:fatherless}

Notice that bottom up evaluation
only requires the child's new subtree,
e.g.\
the subtree it will inherit from is ``father''.
The rest of the code is identical to
the parent it inherits its root node from
(its ``mother'').
Thus it is easy to implement
evaluating the child's fitness 
before the child has been created
(mentioned in Section~\ref{sec:fast_xo} above).
Given everyone in the population's fitness,
it is possible to perform selection
and discover who will mate with whom
and who, even by just bad luck,
will not mate.

In every evolutionary algorithm,
if crossover or mutation is going to be expensive
and we know that the individual is not
going to pass its genetic material on,
we do not have to create it.

Early in an EA run,
i.e.\ before convergence,
the saving can be large.
For example, with tournament selection
and mutation only,
we can save creating on average
$1-1/({\rm tournament\ size})$
of the children.
E.g.\ in a generational EA 
with a diverse population of fitness values,
no elitism, only mutation,
and with tournaments of size~7,
on average 
86\% of the next population will not themselves 
have children and so can be safely ignored.
Even with 100\% two parent crossover and 
a fully converged population,
$1-e^{2}$ $=14\%$ of children will not themselves have children
and again need not be created.

\section{Deep Expressions Hide Errors}
\label{sec:FDP}

\subsection{Information Theory}

Information theory suggests what we have
described for floating point~\cite{langdon:GPEM:gpconv,Langdon:2022:ALJ}
and now integer genetic programming~\cite{langdon:2022:GECCOcomp}
will hold where there aren't side effects to carry
information long distances through program code
or GP trees
(see Figure~\ref{fig:david2_colour},
   page~\pageref{fig:david2_colour}).
Essentially the argument is: 
for a mutation, change, bug, transient error, etc.,
to have an effect it must give rise to a change
of state
and that change of state must propagate from the error
to the program's output~%
\cite{Petke:2021:FSE-IVR}.
However,
without side effects,
the change must pass through a series of nested digital computations
between it and the program's output.
Typically each computation has fewer output states 
than it has inputs states.
(In deterministic programs it can have the same
number of output states as it has input states
but it cannot have more.)
As we saw in Section~\ref{sec:gpinc}
with bottom up evaluation of trees,
if at any point,
the disrupted execution creates the same state
as the original correct execution,
the twin executions remain
locked together,
so that when execution reaches
any output,
the correct and erroneous calculation,
generate identical output.
The error has disappeared%
\footnote{%
Notice failed disruption propagation (FDP) holds in all digital systems,
not just genetic programming.
Any disruption in any deterministic system will tend not
to propagate as it passes through 
intermediate operations.
}.

Indeed so too has any fitness signal
from the mutation.
Without fitness to guide it,
evolution becomes an undirected random walk.

\subsection{Better Testing May Only Help a Little}

For GECCO 2022~\cite{langdon:2022:GECCO2}
we showed if fitness tests are independent of each other 
and equally effective,
then the effectiveness of the whole test set
increases only slowly,
O$(\log n)$,
with increasing number $n$ of fitness tests.
If the tests not equal,
the test set is dominated by the most effective tests
and the weaker ones contribute little.
Of course if we have insight into the code being
tested we may be able to make much better intelligent
choices of test cases values.

\subsection{Will Side Effects Help?}

\begin{figure}
\centerline{\includegraphics[width=\columnwidth]{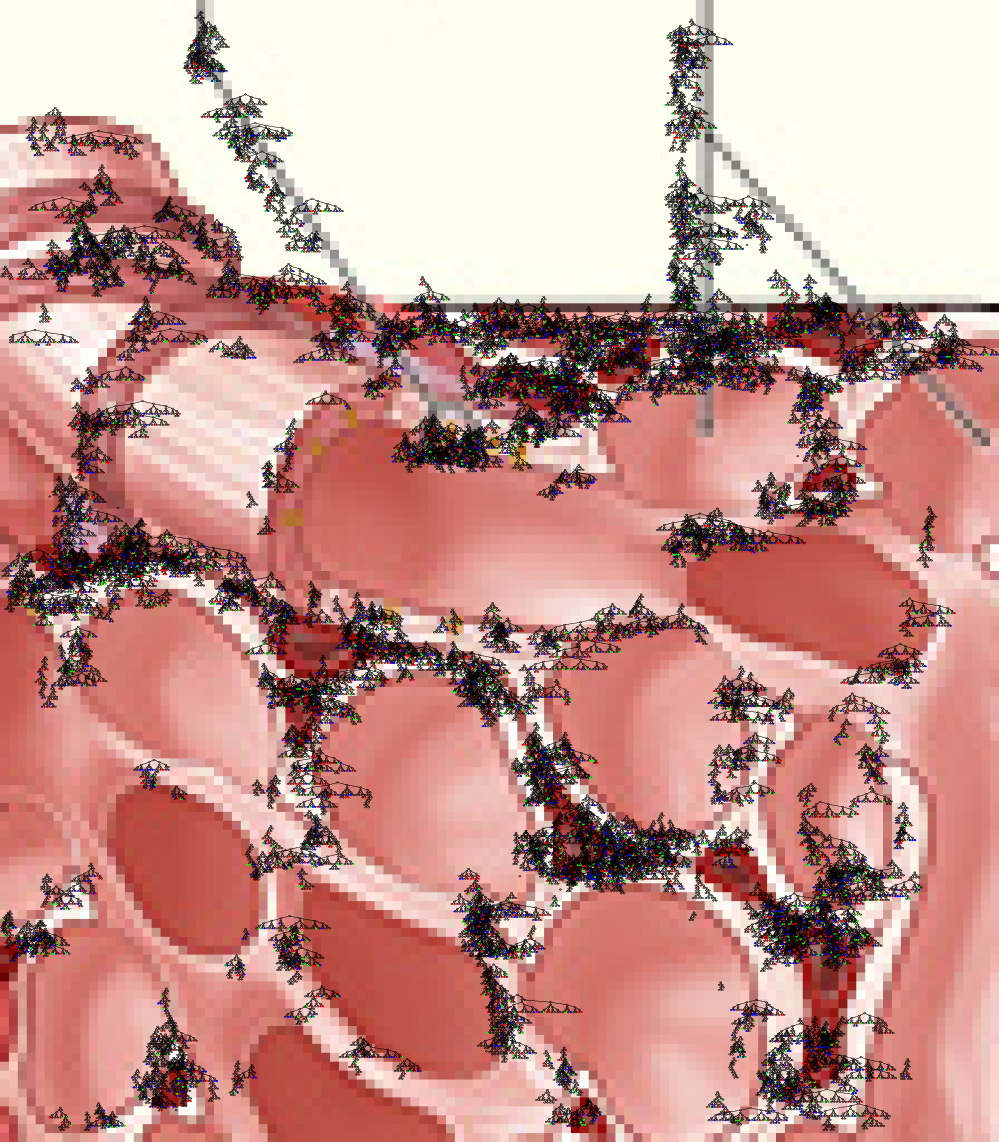}}
\caption{
Proposed
lung like open complex evolving system
composed of 1300 separate shallow GP programs. 
These compute element are placed 
side-by-side to form an open structure.
The gaps promote short cut side effects between functions'
input and outputs and the environment.
From~\cite{langdon:2021:sigevolution}. 
\label{fig:lung_like}
}
\end{figure}

In a recent issue of the
ACM Special Interest Group on Genetic and Evolutionary Computation's
newsletter~\cite{langdon:2021:sigevolution},
I argue that to evolve large complex artifacts of any sort,
we must give up on any hope of evolving large deep structures
but instead limit their effective depth 
by evolving them close to their environment.
Perhaps adapting an open structure
(see Figure~\ref{fig:lung_like},
   page~\pageref{fig:lung_like}),
where a flood tide can bring in new inputs
and and out going ebb tide can move the output of shallow
structures both to the environment and to other parts
of our environmentally embedded evolving mycelium
like program or artifact.

\section{Conclusions}

We have shown a number of speed up techniques
which could be implemented in other evolutionary algorithms.
The C++ code used by GPquick is available
or you may prefer to make your own implementation
of these ideas.

As a few of the citations have shown,
the ideas of distributed parallel running
are widely known in evolutionary algorithms.
However GPU and TPU implementations 
are at present thinner on the ground.
Replicating SETI@home's success
with evolutionary search might also be worth another look.

Similarly if you are interested in using EAs 
on problems with a population of large chromosomes,
some of the ideas for reducing
memory consumption or
speeding up crossover/mutation
might be useful.

Alternatively if you are hoping to run evolution
on tiny ``mote'', embeded, internet of things (IOT)
or ``edge computing'' devices,
these ideas for shoehorning genetic programming~(GP)
into the available memory
may also be interesting.

The vast bulk of GP uses static fitness
test and primitives without side effects
and thus the GPquick implementation
I have sketched could be of interest.

Perhaps the most important result,
due to its universality,
was unlooked for:
as structures get deeper it becomes 
harder to measure the impact of changes.
Although good choices of test cases can help,
simply increasing the number $n$ of test cases
can perhaps only make a slow O($\log(n)$) improvement.
The impossibility of deeply probing
applies to programming in general
not just to evolutionary algorithms.
Notice the prevalence of unit testing
shows this is somewhat tacitly accepted in Software Engineering,
because unit testing places
the test oracle, the check for correct output,
close to the error.
In the case of SBSE, EAs, Artificial Life, etc.,
concealing deep errors, bugs, mutations, etc.,
means it is hard to see if search is proceeding.
Without a fitness delta,
we cannot hope to evolve anything deep.

\subsection*{Acknowledgements}
I would like to thank 
the 
\href{https://dev.heuristiclab.com/trac.fcgi/wiki/EvoSoft}
{EvoSoft 2022} reviewers,
Dagstuhl Seminars 
\href{https://www.dagstuhl.de/en/program/calendar/semhp/?semnr=17191}
{17191} on the theory of randomized heuristics and
\href{https://www.dagstuhl.de/en/program/calendar/semhp/?semnr=18052}
{18052} on
Genetic Improvement of Software~\cite{gi_dagstuhl_2018},
for inspiring conversations
and
the Meta OOPS project.

C++ code is available in
\url{http://www.cs.ucl.ac.uk/staff/W.Langdon/ftp/gp-code/GPinc.tar.gz}

\bibliographystyle{ACM-Reference-Format}

\bibliography{/tmp/gp-bibliography,/tmp/references}

\end{document}